\documentclass{article}

\usepackage{arxiv}

\usepackage[utf8]{inputenc} 
\usepackage[T1]{fontenc}    
\usepackage{hyperref}       
\usepackage{url}            
\usepackage{booktabs}       
\usepackage{amsfonts}       
\usepackage{nicefrac}       
\usepackage{microtype}      
\usepackage{cleveref}       
\usepackage{lipsum}         
\usepackage{graphicx}
\usepackage{natbib}
\usepackage{doi}
\usepackage{multirow}
\usepackage{booktabs}
\usepackage{xcolor}
\usepackage{float}
\usepackage{subcaption}

\title{SiamGPT: Quality-First Fine-Tuning for Stable Thai Text Generation}

\date{}

\newif\ifuniqueAffiliation

\ifuniqueAffiliation 
\author{{xxx yyy} \\
	aaa, Thailand\\
	\texttt{yyy@aaa.co.th} \\
}
\else
\usepackage{authblk}

\author[1]{%
	{Thittipat Pairatsuppawat\thanks{\texttt{thittipat.p@siam.ai}}}%
}
\author[1]{%
	{Abhibhu Tachaapornchai\thanks{\texttt{abhibhu.t@siam.ai}}}%
}
\author[1]{%
	{Paweekorn Kusolsomboon\thanks{\texttt{paweekorn.k@siam.ai}}}%
}
\author[1]{%
	{Chutikan Chaiwong\thanks{\texttt{chutikan.c@siam.ai}}}%
}
\author[2,3]{%
 	{Thodsaporn Chay-intr\thanks{\texttt{t.chayintr@gmail.com}}}%
}
\author[2,3,4]{%
 	{Kobkrit Viriyayudhakorn\thanks{\texttt{kobkrit@aieat.or.th}}}%
}
\author[5,7]{
    {Nongnuch Ketui\thanks{\texttt{nongnuchketui@rmutl.ac.th}}}%
}
\author[6,7]{
    {Aslan B. Wong\thanks{\texttt{aslan.b@sigchi.org}}}%
}

\affil[1]{SIAM.AI}
\affil[2]{iApp Technology Co., Ltd.}
\affil[3]{Intelligent Informatics and Service Innovation Research Center, Thailand}
\affil[4]{Artificial Intelligence Entrepreneur Association of Thailand (AIEAT)}
\affil[5]{Rajamangala University of Technology Lanna Nan, Thailand}
\affil[6]{National Electronics and Computer Technology Center (NECTEC)}
\affil[7]{Artificial Intelligence Association of Thailand (AIAT)}
\fi

\renewcommand{\headeright}{Technical Report}
\renewcommand{\undertitle}{\includegraphics[width=0.12\textwidth]{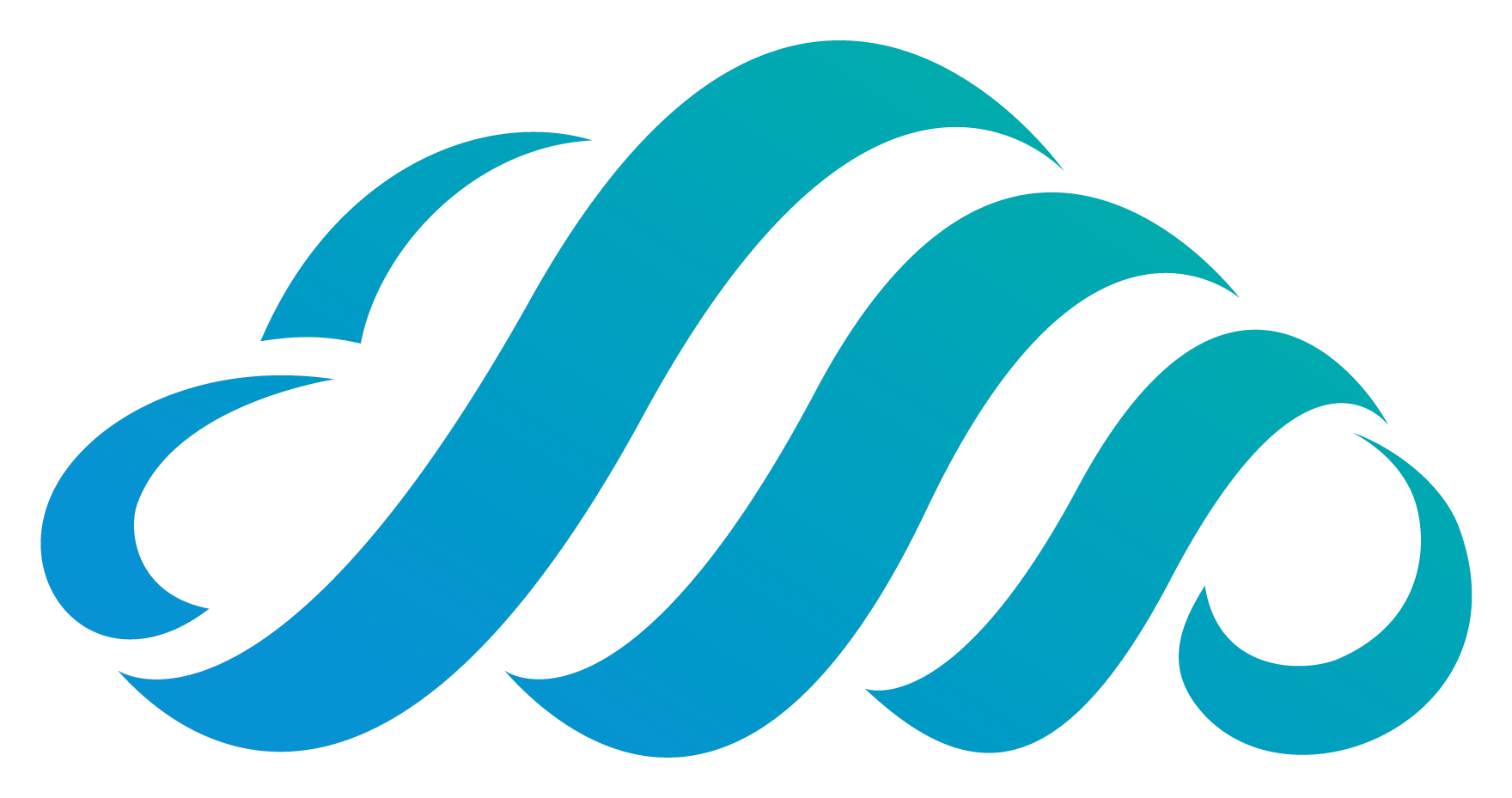}\\[0.5ex]SIAM.AI Cloud\\[0.5ex]Technical Report}
\renewcommand{\shorttitle}{SiamGPT}
\hypersetup{
    pdftitle={SiamGPT: Quality-First Fine-Tuning for Stable Thai Text Generation},
    pdfsubject={cs.CL},
    pdfauthor={Thittipat Pairatsuppawat, Abhibhu Tachaapornchai, Paweekorn Kusolsomboon, Chutikan Chaiwong, Thodsaporn Chay-intr, Kobkrit Viriyayudhakorn, Nongnuch Ketui, Aslan B. Wong},
    pdfkeywords={LLM, Thai},
}

\begin{document}
\maketitle
\begin{abstract}
Open-weights large language models remain difficult to deploy for Thai due to unstable generation under complex instructions, despite strong English performance.
To mitigate these limitations, We present \textbf{\textsc{SiamGPT-32B}}, an open-weights model based on Qwen3-32B, fine-tuned with a \textbf{Quality-First} strategy emphasizing curated supervision over data scale. The fine-tuning pipeline combines high-complexity English instruction data with a Thai-adapted AutoIF framework for instruction and linguistic constraints. Using supervised fine-tuning only, without continual pretraining or corpus expansion, \textsc{SiamGPT-32B} improves instruction adherence, multi-turn robustness, and linguistic stability. Evaluations on the SEA-HELM benchmark show that \textsc{SiamGPT-32B} achieves the strongest overall performance among similar-scale open-weights Thai models, with consistent gains in instruction following, multi-turn dialogue, and natural language understanding.

\textbf{\textsc{SiamGPT-32B:}} \href{https://huggingface.co/siamaids/SiamGPT-32B}{https://huggingface.co/siamaids/SiamGPT-32B \includegraphics[height=1.0em]{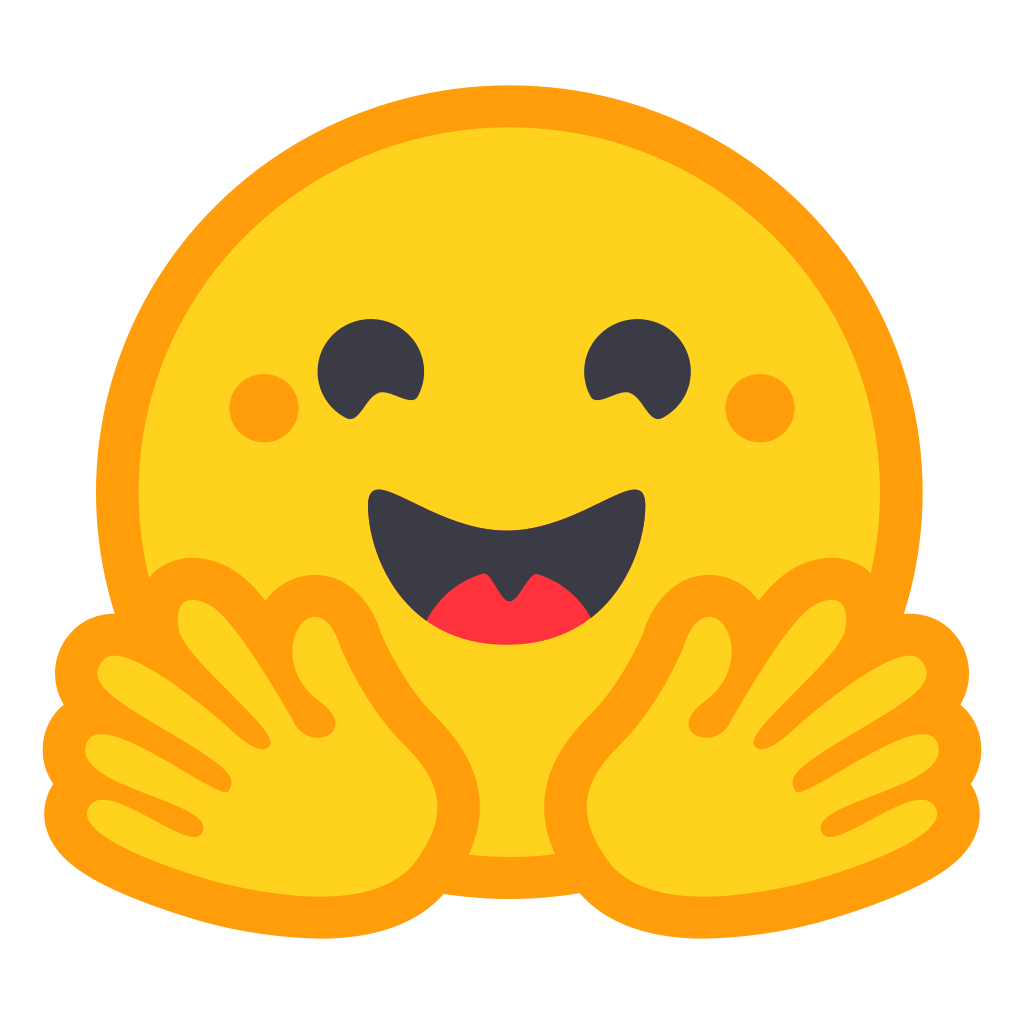}}

\begin{figure}[ht]
    \centering
    \includegraphics[width=0.99\linewidth]{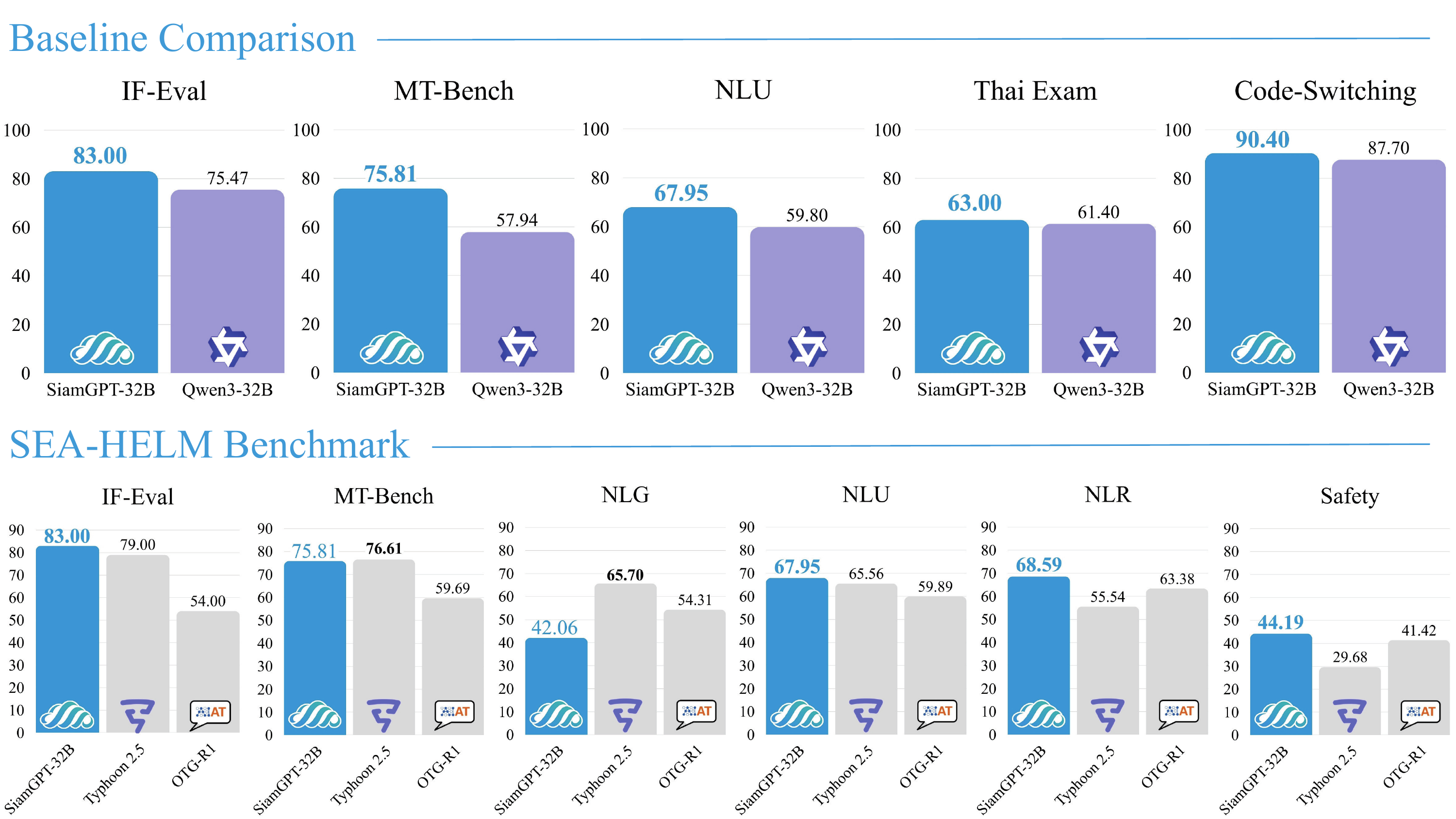}
    \caption{\textsc{SiamGPT-32B} Main Results}
    \label{fig:main-results}
\end{figure}

\end{abstract}

\section{Introduction}
Although open-weights large language models (LLMs) such as Gemma, Qwen, Deepseek, and Mistral \citep{gemmateam2025gemma3technicalreport,yang2025qwen3technicalreport,deepseekai2025deepseekv3technicalreport,mistralai2025magistral} achieve strong performance on English benchmarks without additional fine-tuning, their Thai-language performance, while non-trivial, remains noticeably weaker and more sensitive to instruction complexity without further adaptation.
Consequently, recent Thai-centric efforts such as Typhoon \citep{pipatanakul2023typhoonthailargelanguage,pipatanakul2024typhoon2} and OpenThaiGPT (OTG) \citep{yuenyong2024openthaigpt15,yuenyong2025openthaigpt16r1thaicentric} focus on adapting multilingual base models through various fine-tuning approaches to improve Thai benchmark performance. 

While these efforts substantially improved Thai language understanding and benchmark performance, generation-time issues such as code-switching, instruction sensitivity, and multi-turn inconsistency can still be observed in Thai text generation.
These limitations are particularly problematic in production settings where models act as final response generators within agentic or tool-augmented systems and must reliably synthesize upstream outputs into fluent Thai \citep{plaat2025agenticlargelanguagemodels,zhang2025agenticcontextengineeringevolving,zhang2025deepanalyzeagenticlargelanguage}.

In our preliminary evaluation of \texttt{Qwen3-32B}\footnote{\url{https://huggingface.co/Qwen/Qwen3-32B}}, we observed frequent code-switching, where non-Thai tokens (e.g., Chinese, Hindi, or English) are injected into Thai outputs, often corrupting named entities.
Such failures reduce reliability in user-facing deployments and frequently lead practitioners to rely on proprietary models, such as those provided by OpenAI\footnote{\url{https://openai.com}}, Gemini\footnote{\url{https://gemini.google.com}}, or Claude\footnote{\url{https://claude.ai}}, instead.

To address these generation-time issues, we present \textbf{\textsc{SiamGPT-32B}}, an open-weights, fine-tuned variant of \texttt{Qwen3-32B} optimized for stable Thai generation in multi-turn and instruction-sensitive settings.
The model focuses on improving output stability, instruction following, and multi-turn consistency without relying on continual pretraining or large-scale Thai data collection.

Our approach follows a \textbf{Quality-First} design philosophy \citep{li2024quantityqualityboostingllm} that prioritizes carefully curated supervision over data scale.
Specifically, we fine-tune the model using high-complexity English instruction-following data, focusing on instruction adherence and multi-turn interaction.
We employ an AutoIF framework adapted to incorporate Thai-specific linguistic and formatting constraints, which are enforced through deterministic verification during supervised fine-tuning.
This pipeline produces a compact, high-fidelity training corpus that is used for supervised fine-tuning of \textsc{SiamGPT-32B}, directly targeting generation stability, instruction adherence, and multi-turn consistency.

Experimental results indicate that \textsc{SiamGPT-32B} achieves the strongest overall performance among comparably sized open-weights models, with substantial gains in instruction following, multi-turn dialogue, natural language understanding, and generation stability. These results demonstrate that stable Thai generation can be achieved without continual pretraining or large-scale data expansion, and that carefully curated, constraint-aware supervision is sufficient to address key generation-time failures observed in existing open-weights LLMs.

\section{Approach}
\subsection{Design Goal and Quality-First Strategy}
Our data engineering strategy targets the requirements of a stable Thai \textit{final response generator} for instruction-sensitive and multi-turn settings.
In such deployments, the model must reliably follow formatting constraints, preserve contextual consistency, and produce fluent Thai output without introducing multilingual artifacts.
Rather than relying on noisy, large-scale web crawls, we adopt a \textbf{Quality-First} strategy that prioritizes carefully curated supervision with high reasoning density over data scale.

\subsection{Data Curation Pipeline}\label{section:data-curation-pipeline}

\begin{figure}[h]
    \centering        
    \includegraphics[width=0.99\linewidth]{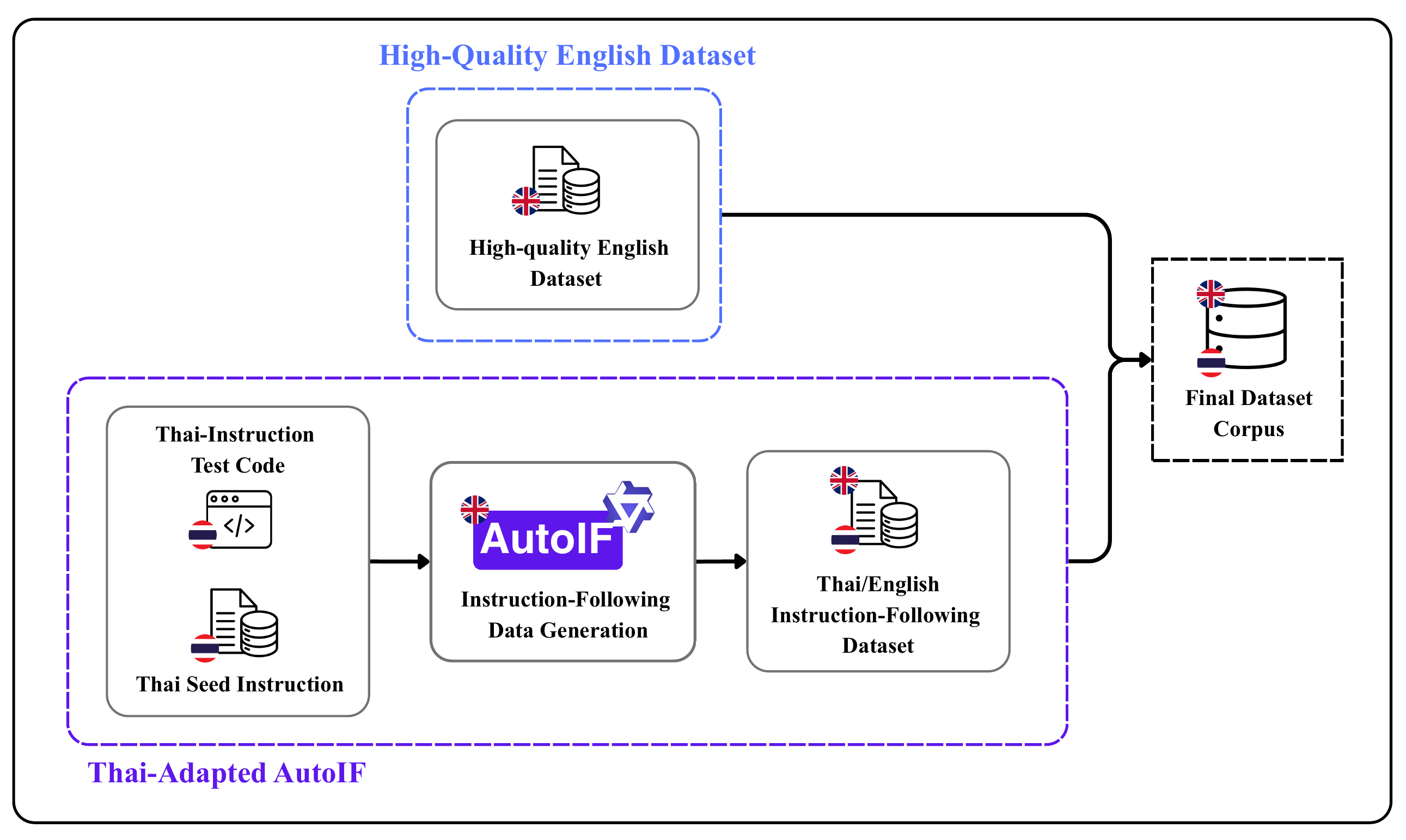}
    \caption{SiamGPT data curation pipeline.
    The pipeline follows a dual-stream design:
    (top) a high-quality English instruction-following data stream that provides diverse reasoning structures and multi-turn conversational patterns for supervised fine-tuning;
    (bottom) a Thai-adapted AutoIF pipeline that enforces instruction-following behavior and Thai-specific linguistic and formatting constraints through deterministic verification.
    Outputs from both streams are merged into a compact, high-fidelity training corpus used for supervised fine-tuning.}
    \label{fig:data-curation}
\end{figure}

To operationalize the Quality-First strategy, we construct a dual-stream data curation pipeline illustrated in Figure~\ref{fig:data-curation}.
The first stream leverages high-quality english instruction-following datasets to transfer reasoning structures and conversational patterns, while mitigating catastrophic forgetting during fine-tuning \citep{luo2025empiricalstudycatastrophicforgetting}.
The second stream enforces strict instruction-following behavior and Thai-specific linguistic and formatting constraints through a Thai-adapted AutoIF framework.
The outputs of both streams are merged into a compact, high-fidelity corpus used for supervised fine-tuning.

\subsubsection{High-quality English Dataset}\label{section:high_quality_english_dataset}

High-quality English instruction-following datasets provide dense supervision for reasoning, dialogue control, and instruction adherence.
In this work, we leverage such datasets to supply complex reasoning structures and multi-turn conversational patterns during supervised fine-tuning.
This supervision helps stabilize instruction-following behavior and mitigates catastrophic forgetting when adapting the model to downstream usage scenarios \citep{luo2025empiricalstudycatastrophicforgetting}.

\subsubsection{Thai-Adapted AutoIF Constraint Enforcement}
While high-quality english dataset transfers reasoning structure, it is insufficient for enforcing strict instruction adherence and Thai orthographic correctness.
To address this gap, we adapt the AutoIF framework to the Thai linguistic context.

AutoIF replaces subjective human annotation with deterministic, programmatic verification by validating model outputs against explicit constraints using executable scripts.
We retain the original English AutoIF seed instructions to preserve universal instruction-following behavior and mitigate catastrophic forgetting \citep{luo2025empiricalstudycatastrophicforgetting}.

In addition, we introduce a custom Thai AutoIF component consisting of 39 manually curated Thai seed instructions.
Following prior evidence that carefully designed, language-specific supervision improves Thai dataset quality and controllability \citep{pipatanakul2024typhoon2}, these instructions target Thai-specific phenomena such as vowel placement rules, consonant usage constraints, and controlled lexical forms. This dual-constraint design preserves high-level reasoning capabilities while enabling precise control over Thai text generation.

\subsection{Final Corpus Construction and Ablation Findings}
We initially explored broad-spectrum training recipes, including large mixed Supervised Fine-Tuning (SFT) corpora, Direct Preference Optimization (DPO), and hybrid SFT+DPO pipelines.
Extensive ablation studies revealed that increasing data volume degraded generation stability and coherence, and that DPO integration yielded marginal gains in instruction-following metrics while causing regressions in other performance dimensions \citep{li2025limrrlscaling}.

Based on these findings, we adopt a minimalist, Pareto-optimal corpus composed of two high-fidelity sources: \texttt{QuixiAI/SystemChat-2.0} and \texttt{AutoIF (English + Thai)}.
This compact corpus prioritizes stability, instruction adherence, and multi-turn consistency over coverage, and forms the sole supervision used to fine-tune \textsc{SiamGPT-32B}.

\subsection{Training Objective and Fine-Tuning Design}
\label{section:fine-tuning-design}

Given the curated corpus, we fine-tune \textsc{SiamGPT-32B} using supervised fine-tuning (SFT) with a standard next-token prediction objective over full instruction--response sequences.
The model is initialized from the \texttt{Qwen3-32B} instruction-tuned checkpoint.

Although SFT is widely used in prior Thai-centric and multilingual adaptation work, our training design deliberately restricts the optimization recipe to SFT on a compact, high-fidelity corpus.
We explicitly avoid continual pretraining, preference optimization, and reinforcement learning.
This restriction isolates the effect of constraint-aware supervision and mitigates instability introduced by reward modeling and large-scale data mixing.

The resulting model is optimized to function as a \emph{final response generator}, prioritizing stable Thai output, strict instruction adherence, and robust multi-turn behavior, rather than planning or tool selection.

\section{Experimental Settings}
This section describes the training corpus, computational infrastructure, optimization settings, and evaluation protocol used to fine-tune and assess \textsc{SiamGPT-32B}.

\subsection{Training Corpus}
The final training corpus consists of two high-fidelity sources totaling approximately \textbf{320,000 instruction--response pairs}, selected through the Quality-First curation process described in Section~\ref{section:data-curation-pipeline}.

\textbf{SystemChat-2.0.}\footnote{\url{https://huggingface.co/datasets/QuixiAI/SystemChat-2.0}}
The dataset contains system-prompted dialogues ranging from 3 to 25 turns per conversation.

\textbf{AutoIF (English + Thai).}\footnote{\url{https://huggingface.co/datasets/siamaids/SiamAI_AutoIF}}
We retained the original 36 English seed instructions from the AutoIF framework and augmented them with 39 manually curated Thai seed instructions targeting Thai-specific orthographic constraints.
Through the AutoIF synthesis and verification pipeline, this produced approximately 180,000 validated instruction--response pairs.

\subsection{Training Framework and Infrastructure}
We initialize training from the \texttt{Qwen3-32B} model. Fine-tuning is conducted using Volcano Engine Reinforcement Learning (VeRL)\citep{Sheng_2025}, a high-performance post-training framework that supports flexible dataflow composition and integrates Fully Sharded Data Parallelism (FSDP) with Flash Attention.

All experiments are run on a distributed cluster of 8 nodes, each equipped with 8 NVIDIA H100 (80GB) GPUs, yielding a total of 64 GPUs.
Training is performed using \texttt{BF16} mixed precision for numerical stability and memory efficiency.

\subsection{Optimization and Hyperparameters}
We perform supervised fine-tuning (SFT) using the high-fidelity corpus described above.
Following the post-training protocol established for \texttt{Qwen2.5} \citep{qwen2025qwen25technicalreport}, we omit learning-rate warmup, as prior work shows that instruction-tuned checkpoints benefit from immediate exposure to the target learning rate rather than gradual warmup.

Training proceeds for 4,096 optimization steps with a global batch size of 512, processing approximately \textbf{2.1 billion tokens} in total (512 $\times$ 4,096 $\times$ 1,000 effective tokens per sample after packing).
With a micro-batch size of 2 per GPU across 64 GPUs, this corresponds to a gradient accumulation factor of 4.
Sequence packing is enabled to maximize token utilization.
The complete optimization and system-level hyperparameters are summarized in Table~\ref{tab:hyperparameters}.

\begin{table}[ht]
\centering
\renewcommand{\arraystretch}{1.2}
\begin{tabular}{lr}
\toprule
\textbf{Parameter} & \textbf{Value} \\
\midrule
Base Model & \texttt{Qwen3-32B} \\
Precision & \texttt{BF16} \\
Max Sequence Length & 8,192 tokens \\
\midrule
Global Batch Size & 512 \\
Micro-Batch Size (per GPU) & 2 \\
Gradient Accumulation & 4 \\
Learning Rate & $1 \times 10^{-5}$ \\
LR Schedule & Cosine (10\% warmup) \\
Optimizer & AdamW \\
Weight Decay & 0.01 \\
Gradient Clipping & 1.0 \\
\midrule
Training Steps & 4,096 \\
Total Tokens & $\sim$2.1B \\
GPUs & 8 nodes $\times$ 8 H100 (80GB) GPUs (64 total) \\
Distributed Training & FSDP2 (full sharding) \\
Attention Kernel & Flash Attention v2 (Liger) \\
Gradient Checkpointing & Enabled \\
Sequence Packing & Enabled \\
\bottomrule
\end{tabular}
\vspace{0.3cm}
\caption{Hyperparameters and system configuration for SFT of \textsc{SiamGPT-32B}. Training completed in approximately 7 hours on a cluster of 8 nodes, each equipped with 8 NVIDIA H100 (80GB) GPUs (64 GPUs total).}
\label{tab:hyperparameters}
\end{table}

\subsection{Evaluation Benchmarks}\label{section:eval}
Model performance is evaluated using the SEA-HELM benchmark suite, which provides standardized evaluation across six competencies: Natural Language Understanding (NLU), Natural Language Generation (NLG), Natural Language Reasoning (NLR), Instruction Following, Safety, and Multi-turn Dialogue. All evaluations were conducted through the official SEA-HELM evaluation pipeline to ensure reproducibility and fair comparison.

\textbf{Instruction Following (SEA-IFEval).}
Instruction adherence is evaluated using SEA-IFEval, a Thai-localized adaptation of IF-Eval \citep{zhou2023ifeval}, translated and culturally adapted by native Thai speakers.
The benchmark measures compliance with explicit formatting and content constraints \citep{susanto2025seahelm}.

\textbf{Multi-Turn Dialogue (SEA-MTBench).}
Multi-turn conversational ability is assessed using Thai-MTBench, developed by VISTEC \citep{payoungkhamdee2024mtbenchthai}, following the LLM-as-a-Judge paradigm \citep{zheng2023mtbench}.
The benchmark consists of 68 Thai-translated prompts and uses GPT-4o (gpt-4o-2024-05-13) as the judge model.

\textbf{NLU/NLG/NLR.}
SEA-HELM’s NLU evaluation includes extractive question answering (XQuAD) and sentiment analysis (Wisesight).
NLG evaluates English--Thai bidirectional translation and abstractive summarization (XLSum), while NLR measures natural language inference (XNLI) and causal reasoning.

\textbf{Thai Exam (ThaiExam).}
Thai-specific knowledge and reasoning are evaluated using ThaiExam \citep{thaiexam2024}, a multiple-choice benchmark covering ONET, TGAT, TPAT-1, A-Level, and IC licensing exams.

\textbf{Safety.} Model safety is evaluated through a toxicity detection task. Models are assessed on their ability to classify Thai text as clean, abusive, or hateful.

\textbf{Code-Switching Score.}
Generation stability is measured as the proportion of Thai-prompted outputs containing only Thai script characters.
Outputs with unexpected non-Thai characters are counted as failures, excluding proper nouns, URLs, and technical identifiers.
This metric captures a common multilingual failure mode and directly reflects Thai-only generation stability.

\section{Results}
We report the performance of \textsc{SiamGPT-32B} on the SEA-HELM benchmark suite following the evaluation protocol described in Section~\ref{section:eval}, with the main results summarized in Figure~\ref{fig:main-results}.

\subsection{Impact of Quality-First Fine-Tuning on Qwen3-32B}
Table~\ref{tab:benchmark_comparison} compares \textsc{SiamGPT-32B} with its base model, Qwen3-32B. Across all benchmarks, \textsc{SiamGPT-32B} outperforms the baseline, demonstrating the effectiveness of Quality-First fine-tuning with constraint-aware supervision.

The largest improvement is observed in \textbf{multi-turn dialogue robustness}. Performance on SEA-MTBench increases from 57.94 to 75.81, indicating that SystemChat-2.0 supervision substantially improves contextual consistency across turns. This shift reflects a transition from single-turn completion behavior to more reliable multi-turn conversational interaction.

\textbf{Instruction following} also improves markedly, with SEA-IFEval increasing from 75.47 to 83.00. This gain reflects the impact of AutoIF-based constraint supervision on formatting and content adherence, which is critical for downstream agentic workflows that require strict output compliance. In parallel, \textbf{generation stability} improves, as the Code-Switching Score increases from 87.70 to 90.40, indicating fewer mixed-script artifacts under Thai prompts.

\textbf{NLU performance} improves from 59.80 to 67.95, suggesting that high-quality instruction supervision enhances language understanding beyond surface-level generation control. Improvements on ThaiExam are more modest, increasing from 61.40 to 63.00, which is consistent with the model’s design focus on stability and controlled generation rather than factual memorization.

Overall, the average score across benchmarks increases from 68.46 to 76.03, confirming that Quality-First fine-tuning yields consistent improvements across stability, instruction following, dialogue robustness, and language understanding.

\begin{table}[ht!]
\centering
\renewcommand{\arraystretch}{1.1}
\begin{tabular}{lcc}
\toprule
\multirow[c]{2}{*}{\textbf{Benchmark}} & \textbf{Qwen3} & \textbf{SiamGPT} \\
\cmidrule(lr){2-2} \cmidrule(lr){3-3}
 & 32B & 32B \\
\midrule
SEA-IFEval (Instruction Following) & 75.47 & \textbf{83.00} (+7.53) \\
SEA-MTBench (Multi-Turn Dialogue) & 57.94 & \textbf{75.81} (+17.87) \\
NLU (QA + Sentiment) & 59.80 & \textbf{67.95} (+8.15) \\
ThaiExam & 61.40 & \textbf{63.00} (+1.60) \\
Code Switching (Stability) & 87.70 & \textbf{90.40} (+2.70) \\
\midrule
\textbf{Average} & 68.46 & \textbf{76.03} (+7.57) \\
\bottomrule
\end{tabular}
\vspace{0.3cm}
\caption{Impact of Quality-First fine-tuning on the Qwen3-32B base model. The table follows the order of the visual comparison: IF-Eval, MT-Bench, NLU, Thai Exam, and Code Switching.}
\label{tab:benchmark_comparison}
\end{table}

\subsection{SEA-HELM Leaderboard Comparison}
We further compare \textsc{SiamGPT-32B} against Thai open-weights models of comparable scale, including Typhoon2.5-Instruct (30B) \citep{pipatanakul2024typhoon2} and OTG-R1 (32B) \citep{yuenyong2025openthaigpt16r1thaicentric}. Results are shown in Table~\ref{tab:sea_helm_leaderboard}.

Across the six SEA-HELM competencies, \textsc{SiamGPT-32B} attains the highest average score, with a mean of 63.59, outperforming Typhoon2.5 and OTG-R1. The model shows its strongest advantages in \textbf{Instruction Following}, \textbf{Natural Language Reasoning}, and \textbf{Safety}, where it consistently leads the comparison. These gains align with the Quality-First fine-tuning design, which emphasizes constraint adherence, reasoning stability, and controlled output behavior.

In \textbf{Natural Language Understanding}, \textsc{SiamGPT-32B} also achieves the highest score, indicating that curated instruction supervision improves comprehension beyond generation-level control. On \textbf{Multi-Turn Dialogue}, Typhoon2.5 slightly outperforms \textsc{SiamGPT-32B}, although the difference is small relative to the substantial margin over OTG-R1. This suggests both models achieve strong conversational performance, with different trade-offs in dialogue modeling.

\textbf{Natural Language Generation} represents the primary area where \textsc{SiamGPT-32B} trails Typhoon2.5. SEA-HELM NLG emphasizes translation fluency and abstractive naturalness, whereas \textsc{SiamGPT-32B} prioritizes stability, formatting control, and reduced code-switching. This difference reflects an intentional design choice, as the model is optimized for reliable final-response generation rather than open-ended generative fluency.

The comparison shows that \textsc{SiamGPT-32B} delivers the strongest aggregate performance among open-weights Thai models in this size class, while exhibiting clear and interpretable trade-offs that reflect its targeted deployment goals.

\begin{table}[ht!]
\centering
\renewcommand{\arraystretch}{1.1}
\begin{tabular}{lccc}
\toprule
\multirow[c]{3}{*}{\textbf{Benchmark}} & \textbf{SiamGPT} & \textbf{Typhoon 2.5} & \textbf{OTG-R1} \\
\cmidrule(lr){2-2} \cmidrule(lr){3-3} \cmidrule(lr){4-4}
 & 32B & 30B & 32B \\
 & Qwen3 & Qwen3 & DeepSeek-R1 \\
\midrule
Instruction Following (SEA-IFEval) & \textbf{83.00} & 79.00 & 54.00 \\
Multi-Turn Dialogue (SEA-MTBench) & 75.81 & \textbf{76.16} & 59.69 \\
NLG (Translation + Summarization) & 42.06 & \textbf{56.70} & 54.31 \\
NLU (QA + Sentiment) & \textbf{67.95} & 65.56 & 59.89 \\
NLR (NLI + Causal Reasoning) & \textbf{68.59} & 55.54 & 65.38 \\
Safety & \textbf{44.19} & 29.68 & 41.42 \\
\midrule
\textbf{Average} & \textbf{63.60} & 60.44 & 55.78 \\
\bottomrule
\end{tabular}
\vspace{0.3cm}
\caption{SEA-HELM benchmark comparison among Thai open-weights models in the 30B–32B class. The rows are ordered to match the visual breakdown: IF-Eval, MT-Bench, NLG, NLU, NLR, and Safety.}
\label{tab:sea_helm_leaderboard}
\end{table}

\section{Discussion}
Despite strong stability and instruction-following performance, \textsc{SiamGPT-32B} inherits several limitations common to LLMs and introduces additional constraints from our training strategy.

\textbf{Factuality and grounding:}
Like other LLMs, \textsc{SiamGPT-32B} can hallucinate or produce factually incorrect claims, especially in open-domain settings or when prompts imply missing context \cite{huang2023hallucination}. This is particularly relevant because our intended deployment is as a \emph{final response synthesizer} in agentic workflows: the quality and truthfulness of the final answer are bounded by the quality of upstream retrieval, tool outputs, and intermediate reasoning traces. Retrieval-augmented generation can reduce unsupported claims by grounding responses in external evidence, but it is not a complete fix and still requires careful system design and verification \cite{lewis2020rag,huang2023hallucination}.

\textbf{Decoding pathologies (repetition and looping):}
Neural text generation is known to exhibit degeneration behaviors, such as repetition, loss of coherence, and looping, depending on the decoding strategy and prompt structure \cite{holtzman2020degeneration}. Consistent with this literature, our internal testing indicates that \textsc{SiamGPT-32B} may loop when used as a standalone creative writer rather than within its intended agentic pipeline. In practice, this suggests deploying SiamGPT with (i) stronger context anchoring, (ii) decoding controls (e.g., repetition penalties / sampling constraints), and (iii) refusal or stop criteria for runaway generations \cite{holtzman2020degeneration,li2020unlikelihood}.

\textbf{Evaluation coverage and metric blind spots:}
Our stability metric operationalizes instability as emitting non-Thai characters under Thai prompts. While this captures a critical production failure mode, it does not fully represent other types of instability (e.g., subtle romanization, punctuation mixing, dialectal variation, or semantic drift). The metric is already designed to allow legitimate mixed tokens and to explain the remaining blind spots (romanization, punctuation mixing, dialect, long-context drift). In addition, widely used instruction and dialog benchmarks capture only slices of real-world behavior: MT-Bench is informative for multi-turn coherence but is also sensitive to judge/model biases, while instruction-following metrics emphasize constraint compliance rather than factual correctness \cite{zheng2023mtbench,zhou2023ifeval}. Broader evaluation should include human preference tests with Thai speakers, stress tests over long contexts, and targeted domain benchmarks for intended deployments. 

\textbf{Security and misuse risks:}
When integrated into tool-using or agentic systems, LLMs are vulnerable to prompt injection and related attacks that attempt to override system intent or exfiltrate sensitive information \cite{owasp2025top10,lliu2024promptinjection}. For safety-critical deployments, we recommend standard secure-by-design measures (least-privilege tool access, input/output filtering, and monitoring) and adversarial testing. Automated red-teaming benchmarks offer a systematic way to assess safety and refusal robustness \cite{mazeika2024harmbench}.

\textbf{Data contamination and evaluation leakage:}
Because our training corpus is derived from publicly available instruction-following datasets, there is a non-zero risk of overlap with downstream evaluation items or closely related paraphrases. While we rely on the official SEA-HELM pipeline for standardized comparison, future work should include systematic decontamination checks (e.g., near-duplicate detection between training data and benchmark prompts) to strengthen the validity of reported gains.

\textbf{Compute and reproducibility constraints:}
Our fine-tuning recipe was executed on a 64$\times$H100 (80GB) cluster, which may limit reproducibility. We therefore plan to provide lighter-weight training variants (e.g., parameter-efficient fine-tuning configurations and smaller model checkpoints) and detailed training scripts to broaden accessibility.

\section{Conclusion and Future Work}
\subsection{Conclusion}
This report introduced \textsc{SiamGPT-32B}, a fine-tuned variant of Qwen3-32B designed to solve a production-critical failure mode for Thai: multilingual interference that manifests as mixed-script or code-switching artifacts in otherwise Thai responses. Our core claim is that a \emph{minimal, curated} supervised fine-tuning recipe can produce a reliable Thai response generator when the data is deliberately selected for control and stability rather than sheer scale. We implement this via a Quality-First high-quality instruction corpus paired with Thai-adapted AutoIF constraints \cite{dong2024autoif}, yielding a compact but high-leverage training set.

Empirically, \textsc{SiamGPT-32B} shows consistent improvements over the Qwen3-32B baseline across the pillars most relevant to agentic deployments: (i) improved stability on a code-switching test (87.70 $\rightarrow$ 90.40), (ii) stronger instruction adherence on IF-Eval \cite{zhou2023ifeval} (75.47 $\rightarrow$ 83.00), and (iii) substantially better multi-turn dialog control on MT-Bench \cite{zheng2023mtbench} (57.94 $\rightarrow$ 75.81). On the SEA-HELM Thai leaderboard \cite{susanto2025seahelm,seahelmleaderboard}, SiamGPT achieves the highest overall score among open-weight Thai models in the 30B-32B class, while also achieving strong instruction-following and multi-turn performance.

However, the SEA-HELM breakdown also highlights an important tradeoff. Despite the strongest overall score, SIAMGPT-32B underperforms compared to leading peers in the NLG (Translation) category \cite{susanto2025seahelm}, plausibly
because our supervision emphasizes controllability (instruction adherence and stability) which may negatively impact native natural language generation scores.

\subsection{Future Work}
At present, \textsc{SiamGPT-32B} is optimized as a dependent node that relies on upstream agentic workflows to supply grounding context. Our next iteration aims to transition toward a standalone \emph{Thai Native Expert} while preserving the stability and instruction-following gains that make SiamGPT effective in production settings. Key directions include:

\textbf{(1) Internalize Thai cultural knowledge in the weights:}
We will expand intrinsic Thai cultural and geographic knowledge (e.g., history, national holidays, anthem, provinces) to reduce reliance on external retrieval for common Thai-local queries and to better match Thai-native expectations in everyday dialog.

\textbf{(2) Embed domain expertise for high-value applications:}
We will integrate specialized Thai-domain datasets (e.g., Thai tourism and local navigation) directly into training so the model can answer domain questions with fewer hallucinations and less dependence on upstream systems.

\textbf{(3) Close the NLG gap without sacrificing stability:}
Given the observed SEA-HELM NLG weakness \cite{susanto2025seahelm}, we will explicitly target Thai naturalness and generation quality while retaining deterministic constraint enforcement. Practically, this suggests adding Thai-native conversational and writing supervision and testing multi-objective training that treats stability as a hard constraint rather than a soft preference.

\textbf{(4) Make corpus decisions evidence-driven via ablations:}
Our internal corpus experiments indicate that simply increasing data breadth degraded stability, and that DPO improved instruction-following but regressed other pillars \cite{rafailov2023dpo}. In the next version, we will formalize these findings with publishable ablation tables and expand them into controlled studies (e.g., which data sources or objectives help NLG the most while preserving stability).

\textbf{(5) Expand evaluation to deployment-style stress tests:}
We will extend evaluation beyond benchmark aggregates to include long-context multi-turn runs, mixed-domain prompts, and “acceptable code-switching” regimes (e.g., proper nouns, URLs, technical identifiers), ensuring the stability metric reflects real user-facing requirements.




\bibliographystyle{unsrtnat}
\bibliography{references}  






\end{document}